# Synthesis of Near-regular Natural Textures

Dr. V. Asha
Dept. of Master of Computer Applications
New Horizon College of Engineering
Bangalore, Karnataka, INDIA
v_asha@live.com

**ABSTRACT**

Texture synthesis is widely used in the field of computer graphics, vision, and image processing. In the present paper, a texture synthesis algorithm is proposed for near-regular natural textures with the help of a representative periodic pattern extracted from the input textures using distance matching function. Local texture statistics is then analyzed against global texture statistics for non-overlapping windows of size same as periodic pattern size and a representative periodic pattern is extracted from the image and used for texture synthesis, while preserving the global regularity and visual appearance. Validation of the algorithm based on experiments with synthetic textures whose periodic pattern sizes are known and containing camouflages / defects proves the strength of the algorithm for texture synthesis and its application in detection of camouflages / defects in textures.

## 1. INTRODUCTION

Texture is an important cue in human and machine vision [1]. Almost all real world objects are textured in nature. Traditionally, textures can be broadly classified as either structural (regular) or stochastic [2]. Most of the real world textures fall between regular and stochastic classes. A structural texture is the one that is characterized by a set of primitives called texons or texels and placement rules like a brick wall texture generated by tiling of bricks in a structured fashion. A perfectly regular texture has periodic patterns, where the color / intensity and shape of all texture elements are repeating in equal intervals along two linearly independent directions [3, 4]. On the other hand, a stochastic texture like grass or sand does not contain explicit primitives.

Near-regular textures can be viewed as statistical departures of regular textures along different dimensions. Near-regular textures, according to Liu et al., [3–5], can be of Type-0 (Regular in Geometry and Color) or Type-I (Regular in Geometry and Irregular in Color) or Type-II (Irregular in Geometry and Regular in Color) or Type-III (Irregular in Geometry and Color).

A texture synthesis algorithm basically starts from a fundamental unit and attempts to generate a complete texture whose appearance is similar to that of the fundamental unit. Primary advantages of texture synthesis include the ability to generate large and/or tilable textures from a small unit directly.

Texture analysis and synthesis algorithms are very important in the field of computer graphics, vision, and image processing. Textures have long been used to decorate object surfaces in computer rendered images. Textures can be synthesized based on mathematical functions (procedural texture synthesis) [6–8] or using an input texture fragment (sample-based texture synthesis) [4, 9 – 11]. Image synthesis or scene generation can be approached from two somewhat different directions, namely, qualitative perceptual-cue approach and quantitative perceptual-cue approach [12]. The goal of the qualitative perceptual-cue approach is to develop synthetic images that have the appearance of a particular type of textured surface, while capturing the appearance quality and invoking a realistic surface appearance to the observer. Examples of this kind of synthesis are the approaches used in video games and animated films. The objective of the quantitative perceptual-cue approach is to generate texture fields that are based on quantitative, physical models underlying important perceptual cues or physical properties. The ability to derive a texture field that has a prescribed quantitative characterization of the underlying physical and perceptual models is considered to be a fundamental concept in developing these algorithms. Such type of algorithm is used in infrared and visible scene generation for design and evaluation of target recognition systems [13, 14].

In machine vision, the ability to discriminate, segment, or classify textures is essential for tasks such as automatic inspection in textile or ceramic / tile industries. As far as natural texture images are concerned, these have high frequency contents. Hence, natural textures are hard to compress and compact representation has to be achieved using a good texture analysis algorithm [10]. Also, creating a robust and general texture synthesis algorithm has been proven to be very much difficult [15]. It is observed that the combination of regularity and randomness of near-regular textures challenges some state of the art texture synthesis algorithms and many general purpose synthesis algorithms fail to preserve the structural regularity in near-periodic textures [3, 4]. Extraction of a representative tile from near-periodic textures with the help of image-editing software manually is very much difficult and time-consuming. Also, the junction between tiles is often not smooth and the appearance is not pleasing, even after multiple attempts [16]. Thus, the development of algorithm which can yield good texture-synthesis remains an open area of research.

As many textures can be considered as combination of periodic repetition of basic elements or building blocks called texels or texture elements [17], in this work, a simple method is presented for synthesis of near-regular natural textures with the help of a representative periodic pattern extracted from the image, while preserving global regularity and aesthetic appearance. This texel is representative in the sense that it alone can be used to generate an image that is faithful to the original image, just by tiling. The idea is to determine the row and column periodicities of the input texture through enhanced version of Distance Matching Function (DMF) proposed in [18] and then to extract a representative periodic pattern based on visual appearance. Once DMFs are computed, textural periodicities and hence periodic pattern size can be determined. For each periodic pattern/ block and for the entire image, first order statistical properties can be evaluated. The first order statistics is based on probability of occurrence of a gray level $r_k$ in an image given by

$$p(r_k) = \frac{n_k}{n}, \; k = 0, 1, 2, \ldots, L-1$$

where, n is the total number of pixels, $n_k$ is the number of pixels that have gray level $r_k$, and L is the total number of gray levels in the image. The properties, mean ($\bar{r}$), variance ($\mu_2$), skewness ($\mu_3$), kurtosis ($\mu_4$), energy ($E_1$), and entropy ($S_1$) can be calculated as given below:

$$\bar{r} = \sum_{k=0}^{L-1} r_k p(r_k)$$

$$\mu_2 = \sum_{k=0}^{L-1} (\bar{r} - r_k)^2 p(r_k)$$

$$\mu_3 = \sum_{k=0}^{L-1} (\bar{r} - r_k)^3 p(r_k)$$

$$\mu_4 = \sum_{k=0}^{L-1} (\bar{r} - r_k)^4 p(r_k)$$

$$E_1 = \sum_{k=0}^{L-1} [p(r_k)]^2$$

$$S_1 = \sum_{k=0}^{L-1} p(r_k) \log_2 [p(r_k)]$$

Synthesis of texture can be made based on analysis of local statistics of each periodic block against global statistics of the input image.

## 2. ALGORITHM, EXPERIMENTS AND RESULTS

In order to extract the periodic pattern which is required for synthesis based on both statistics and visual appearance, it is assumed that the effect of non-uniform or distorted or defective periodic patterns on the global texture properties is negligible. However, the local texture properties of the defective periodic patterns vary considerably with the global statistical properties. Considering this fact, the algorithm for synthesis of natural textures is developed with the following steps:

(i) Determination of row and column periodicities and hence the periodic pattern size
(ii) Determination of the global statistics of the input texture and overlaying an imaginary grid to divide the entire image into blocks each of size same as periodic pattern size
(iii) Computation of local statistics of each periodic pattern and extraction of a representative periodic pattern from the image based on local statistics of each block against global statistics
(iv) Tiling of the representative periodic pattern to form synthesized counterpart of the input image, while preserving the regularity and visual appearance

In order to illustrate the present method of texture synthesis, near-regular natural textures were tested. Figure 1 shows test image-1.

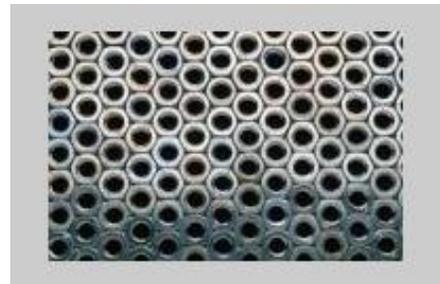

Fig.1: Test image-1.

By fixing a 10% deviation of local statistics against global statistics, totally nine periodic patterns were extracted from the test image using periodicities determined based on [18] and considering visual appearance, a representative periodic pattern is used for texture synthesis. The synthesized texture thus generated using the representative periodic pattern is shown in Fig. 2.

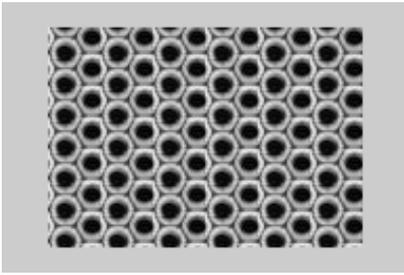

Fig.2: Result of texture synthesis for the test image – 1.

Test image-2 is the texture D20 from Brodatz album and is shown in Fig. 3.

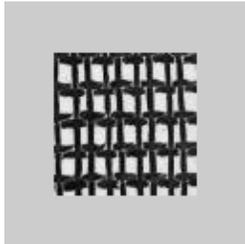

Fig.3: Test image-2 (D20 from Brodatz album).

Here also, based on 10% deviation of local statistics against global statistics, totally twelve periodic patterns were extracted using periodicities determined and considering visual appearance, a representative periodic pattern is used for texture synthesis. The synthesized texture thus generated using the representative periodic pattern is shown in Fig. 4.

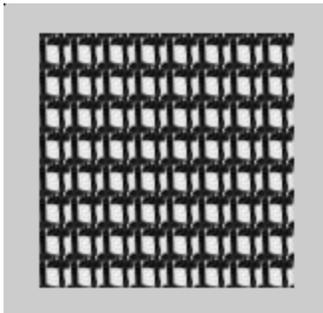

Fig.4: Result of texture synthesis for the test image – 2.

## 3. VALIDATION OF THE ALGORITHM

To validate the present method of synthesis, synthetic images used in our previous work [19] and whose periodic pattern sizes are known were tested. Figure 5 shows a test imageconsisting of cluttered 'U' camouflaged in 'V' background.

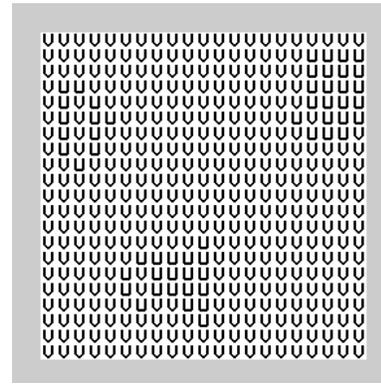

Fig. 5: Test image-3 consisting of cluttered 'U' camouflaged in 'V' background.

The image being synthetic, within 2% deviation of local statistics against global statistics, texture synthesis has been achieved with periodic patterns that are pure background. The synthesized texture is shown in Fig. 6. Periodic patterns which are not selected for synthesis are nothing but camouflages as shown highlighted in Fig. 7.

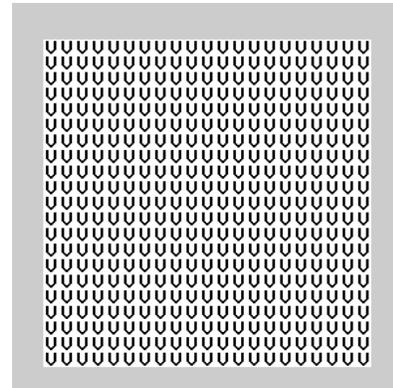

Fig. 6: Result of texture synthesis for the test image – 3.

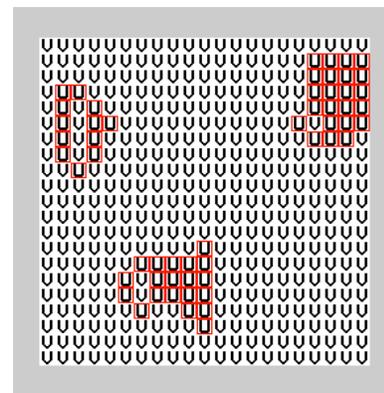

Fig.7: Image highlighting boundary of camouflages of the test image – 3.

Another synthetic image with defective patterns (Fig. 8) was also tested. Here also, within 2% deviation of local statistics against global statistics, texture synthesis has been achieved with periodic patterns that are not defective.

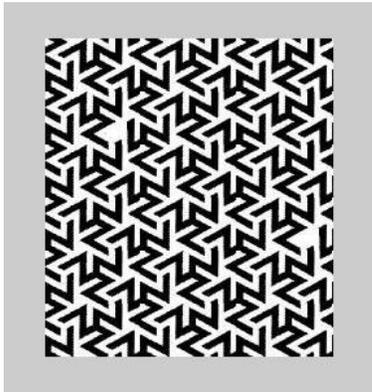

Fig. 8: Test image-4 containing defects.

The result of synthesis is shown in Fig. 9 and the defective periodic patterns which are not used for synthesis are shown highlighted in Fig. 10.

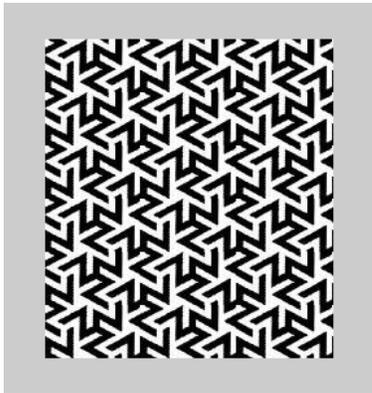

Fig. 9: Result of texture synthesis for the test image – 4.

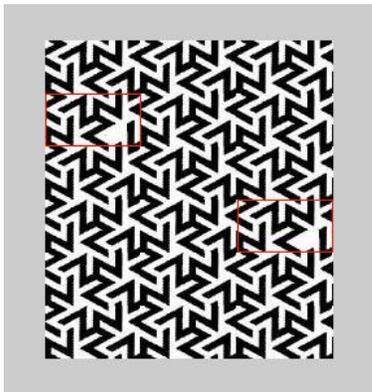

Fig.10: Image highlighting boundary of defective periodic patterns of test image – 4.

## 4. CONCLUSIONS

The algorithm proposed can help in extraction of representative periodic patterns which may be of square or rectangular size from near-regular natural textures and texture synthesis based on analysis of local texture statistics against global texture statistics, while preserving the global regularity and visual appearance. Validation of the algorithm based on experiments with synthetic textures with periodic patterns of known size and containing camouflages/defects has proved the strength of the algorithm for texture synthesis and detection of defects / camouflages.